\documentclass[10pt,twocolumn,letterpaper]{article}

\usepackage{cvpr}              

\usepackage{graphicx}
\usepackage{amsmath}
\usepackage{amssymb}
\usepackage{booktabs}

\usepackage[accsupp]{axessibility}  
\usepackage{algorithm}
\usepackage{algorithmic}
\usepackage{amsfonts,amssymb} 
\usepackage{amsmath}
\usepackage{color}
\usepackage{multirow}
\usepackage{pythonhighlight}
\usepackage{setspace}

\usepackage[pagebackref,breaklinks,colorlinks]{hyperref}

\usepackage[capitalize]{cleveref}
\crefname{section}{Sec.}{Secs.}
\Crefname{section}{Section}{Sections}
\Crefname{table}{Table}{Tables}
\crefname{table}{Tab.}{Tabs.}

\def\cvprPaperID{192} 
\def\confName{CVPR}
\def\confYear{2023}

\begin{document}

\title{SimpleNet: A Simple Network for Image Anomaly Detection and Localization}

\author{Zhikang~Liu$^{1}$\qquad~Yiming~Zhou$^{2}$\qquad~Yuansheng~Xu$^{2}$\qquad~Zilei~Wang$^{1}$\thanks{Corresponding author}\\
		Department of Automation, University of Science and Technology of China$^{1}$ \\
            Meka Technology Co.,Ltd$^{2}$\\
		{\tt\small lzk@mail.ustc.edu.cn} \quad
		{\tt\small zhouyiming.donal@gmail.com} \quad
            {\tt\small xys-tc@hotmail.com} \quad
		{\tt\small zlwang@ustc.edu.cn} 
	}

\maketitle

\begin{figure}[t]
	\centering
	\includegraphics[width=0.46\textwidth]{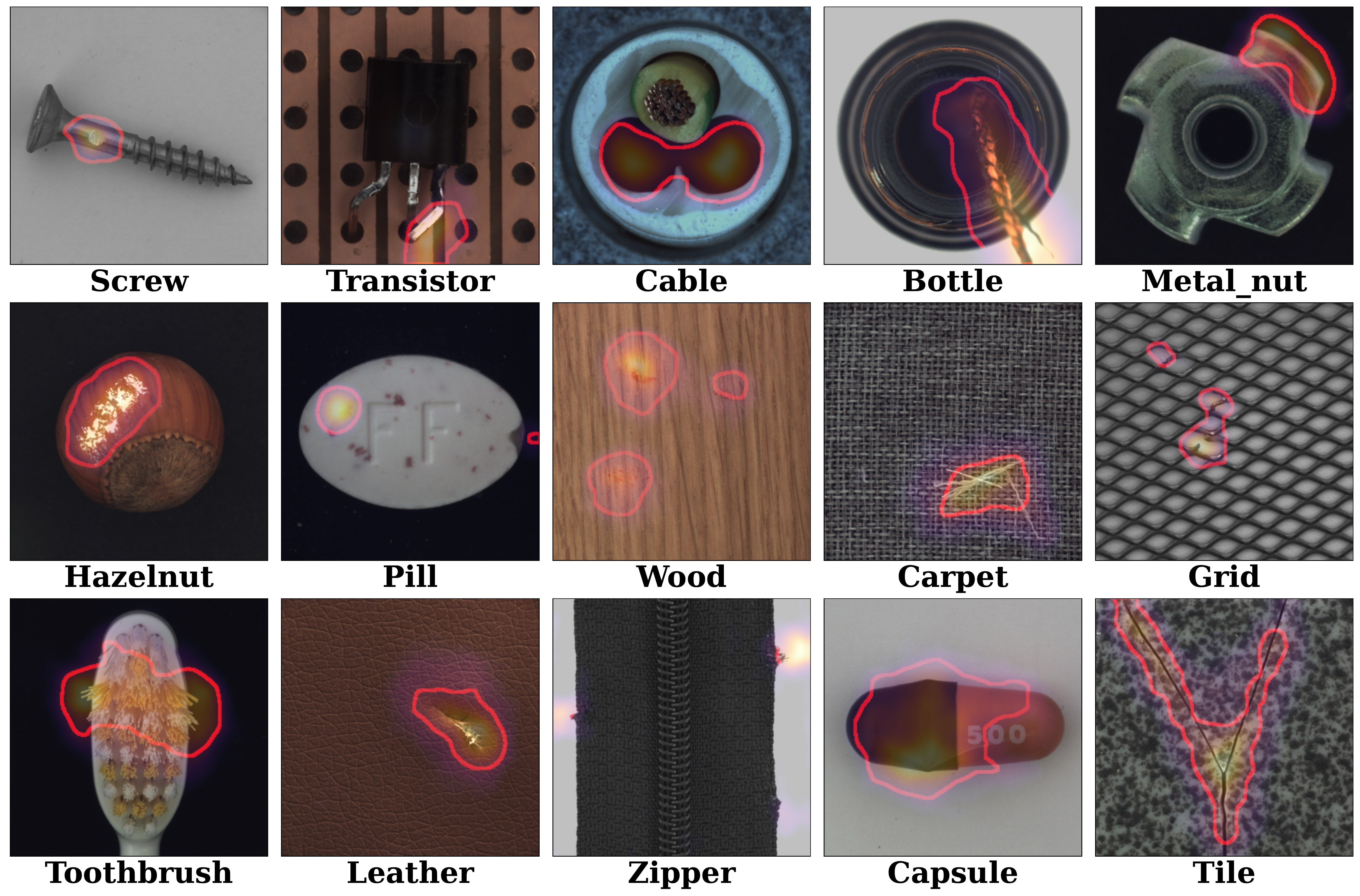} 
	\caption{Visualization of samples in MVTec AD. The produced anomaly maps superimposed on the images. Anomaly region of high anomaly score is colored with \textcolor{orange}{orange}. The \textcolor{red}{red} boundary denotes contours of actual segmentation maps for anomalies.}
	\label{Fig1}
\end{figure}

\begin{abstract}

We propose a simple and application-friendly network (called SimpleNet) for detecting and localizing anomalies. SimpleNet consists of four components: (1) a pre-trained Feature Extractor that generates local features, (2) a shallow Feature Adapter that transfers local features towards target domain, (3) a simple Anomaly Feature Generator that counterfeits anomaly features by adding Gaussian noise to normal features, and (4) a binary Anomaly Discriminator that distinguishes anomaly features from normal features. During inference, the Anomaly Feature Generator would be discarded. 
Our approach is based on three intuitions. First, transforming pre-trained features to target-oriented features helps avoid domain bias. Second, generating synthetic anomalies in feature space is more effective, as defects may not have much commonality in the image space. Third, a simple discriminator is much efficient and practical. In spite of simplicity, SimpleNet outperforms previous methods quantitatively and qualitatively. 
On the MVTec AD benchmark, SimpleNet achieves an anomaly detection AUROC of $99.6 \%$, 
reducing the error by 55.5\% compared to the next best performing model.
Furthermore, SimpleNet is faster than existing methods, with a high frame rate of 77 FPS on a 3080ti GPU.  
Additionally, SimpleNet demonstrates significant improvements in performance on the One-Class Novelty Detection task. Code: \url{https://github.com/DonaldRR/SimpleNet}.
\end{abstract}

\begin{figure}[t]
	\centering
	\includegraphics[width=0.46\textwidth]{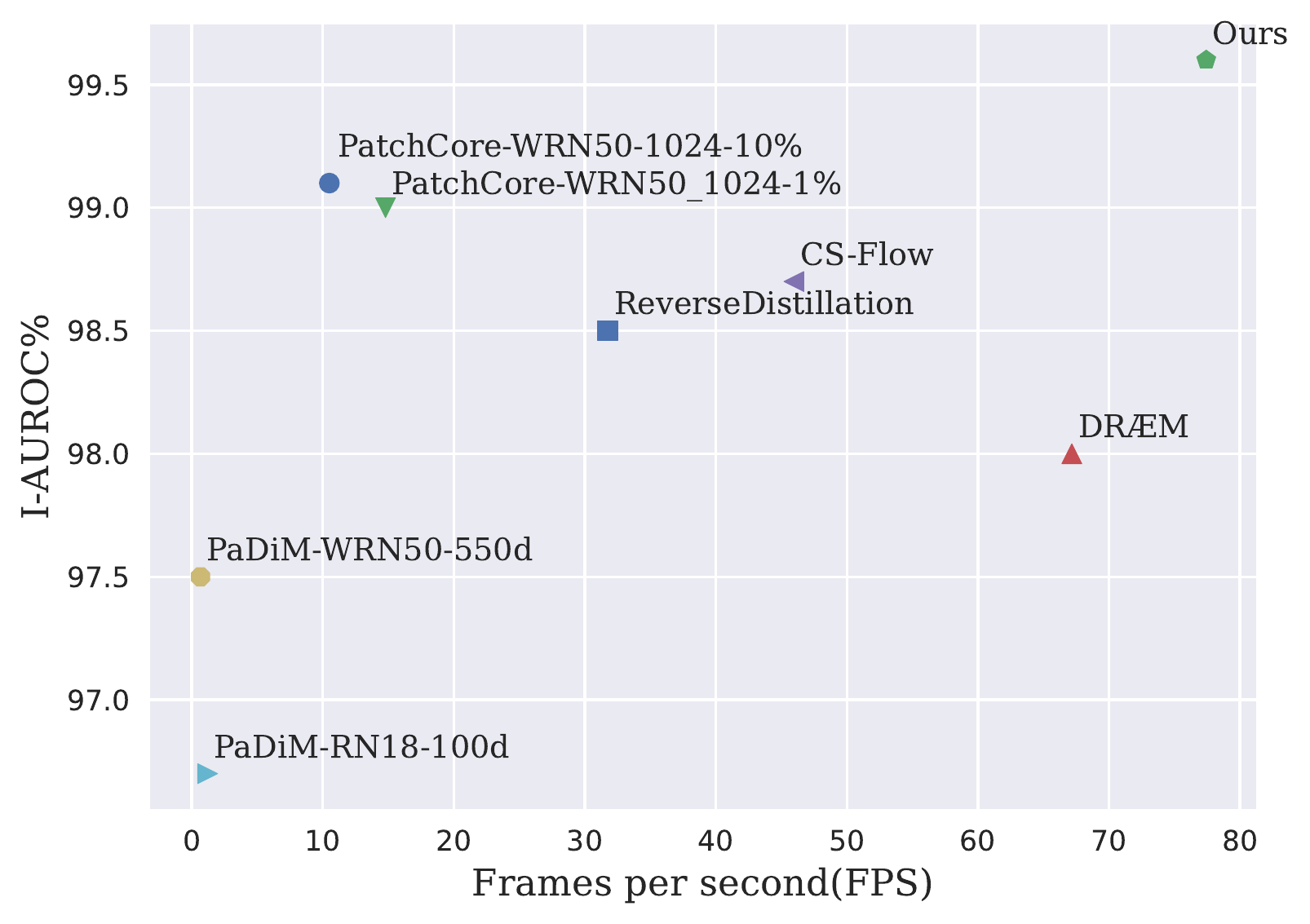} 
	\caption{Inference speed (FPS) versus I-AUROC on MVTec AD benchmark. SimpleNet outperforms all previous methods on both accuracy and efficiency by a large margin.}
	\label{Fig:speed}
\end{figure}

\section{Introduction}

Image anomaly detection and localization task aims to identify abnormal images and locate abnormal subregions. The technique to detect the various anomalies of interest has a broad set of applications in industrial inspection~\cite{bergmann2019mvtec, defard2021padim}. In industrial scenarios, anomaly detection and localization is especially hard, as abnormal samples are scarce and anomalies can vary from subtle changes such as thin scratches to large structural defects, \eg missing parts. Some examples from the MVTec AD benchmark~\cite{bergmann2019mvtec} along with results from our proposed method are shown in Figure~\ref{Fig1}. This situation prohibits the supervised methods from approaching.

Current approaches address this problem in an unsupervised manner, where only normal samples are used during the training process. The reconstruction-based methods~\cite{gong2019memorizing,zavrtanik2021reconstruction,ristea2022self}, synthesizing-based methods~\cite{zavrtanik2021draem, li2021cutpaste}, and embedding-based methods~\cite{defard2021padim, rudolph2022fully, roth2022towards} are three main trends for tackling this problem. The reconstruction-based methods such as~\cite{zavrtanik2021reconstruction,ristea2022self} assume that a deep network trained with only normal data cannot accurately reconstruct anomalous regions. The pixel-wise reconstruction errors are taken as anomaly scores for anomaly localization. However, this assumption may not always hold, and sometimes a network can "generalize" so well that it can also reconstruct the abnormal inputs well, leading to misdetection~\cite{gong2019memorizing, perera2019ocgan}.
The synthesizing-based methods~\cite{zavrtanik2021draem, li2021cutpaste}  estimate the decision boundary between the normal and anomalous by training on synthetic anomalies generated on anomaly-free images. However, the synthesized images are not realistic enough. Features from synthetic data might stray far from the normal features, training with such negative samples could result in a loosely bounded normal feature space, meaning indistinct defects could be included in in-distribution feature space. 

Recently, the embedding-based methods~\cite{defard2021padim, rudolph2022fully, deng2022anomaly,  roth2022towards} achieve state-of-the-art performance. These methods use ImageNet pre-trained convolutional neural networks (CNN) to extract generalized normal features. Then a statistical algorithm such as multivariate Gaussian distribution~\cite{defard2021padim}, normalizing flow~\cite{rudolph2022fully}, and memory bank~\cite{roth2022towards} is adopted to embed normal feature distribution. Anomalies are detected by comparing the input features with the learned distribution or the memorized features. However, industrial images generally have a different distribution from ImageNet. Directly using these biased features may cause mismatch problems. Moreover, the statistical algorithms always suffer from high computational complexity or high memory consumption.

\begin{figure*}[t]
	\centering
	\includegraphics[width=0.8\textwidth]{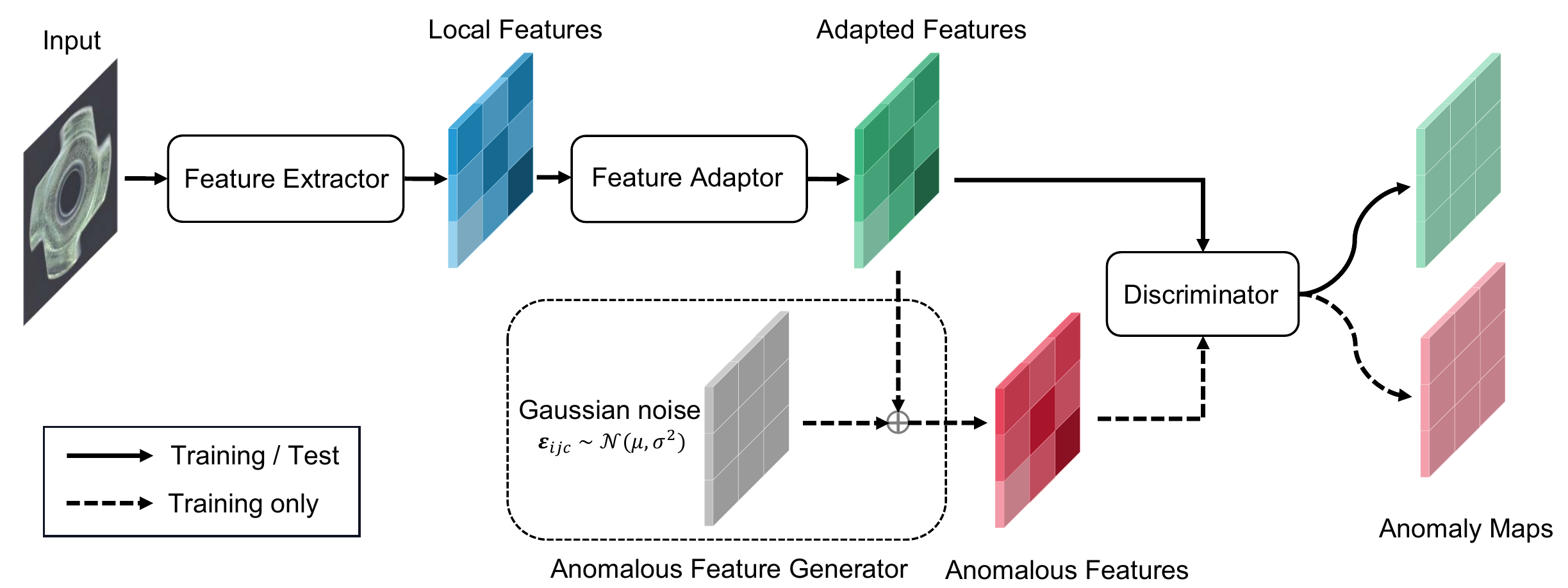} 
	\caption{Overview of the proposed SimpleNet. In the training phase, nominal samples are fed into a pre-trained \textit{Feature Extractor} to get local features. Then, a \textit{Feature Adaptor} is utilized to adapt pre-trained features into the target domain. Anomalous features are synthesized by adding Gaussian noise to the adapted features. The adapted features and the anomalous features and used as positive and negative samples respectively to train the final \textit{Discriminator}. The \textit{Anomalous Feature Generator} is removed at inference.}
	\label{Fig2}
\end{figure*}

To mitigate the aforementioned issues, we propose a novel anomaly detection and localization network, called SimpleNet. SimpleNet takes advantage of the synthesizing-based and the embedding-based manners, and makes several improvements. First, instead of directly using pre-trained features, we propose to use a feature adaptor to produce target-oriented features which reduce domain bias. Second, instead of directly synthesizing anomalies on the images, we propose to generate anomalous features by posing noise to normal features in feature space. We argue that with a properly calibrated scale of the noise, a closely bounded normal feature space can be obtained. Third, we simplify the anomalous detection procedure by training a simple discriminator, which is much more computational efficient than the complex statistical algorithms adopted by the aforementioned embedding-based methods. Specifically, SimpleNet makes use of a pre-trained backbone for normal feature extraction followed by a feature adapter to transfer the feature into the target domain. Then, anomaly features are simply generated by adding Gaussian noise to the adapted normal features. A simple discriminator consisting of a few layers of MLP is trained on these features to discriminate anomalies.

SimpleNet is easy to train and apply, with outstanding performance and inference speed. The proposed SimpleNet, based on a widely used WideResnet50 backbone, achieves 99.6 \% AUROC on MVTec AD while running at 77 fps, surpassing the previous best-published anomaly detection methods on both accuracy and efficiency, see Figure~\ref{Fig:speed}. We further introduce SimpleNet to the task of One-Class Novelty Detection to show its generality. These advantages make SimpleNet bridge the gap between academic research and industrial application. Code will be publicly available.

\section{Related Work}
Anomaly detection and localization methods can be mainly categorized into three types, \ie, the reconstruction-based methods, the synthesizing-based methods, and the embedding-based methods. 

\textbf{Reconstruction-based methods} hold the insight that anomalous image regions should not be able to be properly reconstructed since they do not exist in the training samples. Some methods~\cite{gong2019memorizing} utilize generative models such as auto-encoders and generative adversarial networks~\cite{goodfellow2014generative} to encode and reconstruct normal data. Other methods~\cite{haselmann2018anomaly, zavrtanik2021reconstruction, ristea2022self} frame anomaly detection as an inpainting problem, where patches from images are masked randomly. Then, neural networks are utilized to predict the erased information. Integrating structural similarity index (SSIM)~\cite{wang2004image} loss function is widely used in training. An anomaly map is generated as pixel-wise difference between the input image and its reconstructed image. However, if anomalies share common compositional patterns (e.g. local edges) with the normal training data or the decoder is "too strong" for decoding some abnormal encodings well, the anomalies in images are likely to be reconstructed well~\cite{zavrtanik2021reconstruction}.

\textbf{Synthesizing-based methods} typically synthesize anomalies on anomaly-free images. DRÆM~\cite{zavrtanik2021draem} proposes a network that is discriminatively trained in an end-to-end manner on synthetically generated just-out-of-distribution patterns. CutPaste~\cite{li2021cutpaste} proposes a simple strategy to generate synthetic anomalies for anomaly detection that cuts an image patch and pastes at a random location of a large image. A CNN is trained to distinguish images from normal and augmented data distributions. However, the appearance of the synthetic anomalies does not closely match the real anomalies'. In practice, as defects are various and unpredictable, generating an anomaly set that includes all outliers is impossible. Instead of synthesizing anomalies on images, with the proposed SimpleNet, negative samples are synthesized in the feature space.

\textbf{Embedding-based methods} achieve state-of-the-art performance recently. These methods embed normal features into a compressed space. The anomalous features are far from the normal clusters in the embedding space. Typical methods~\cite{defard2021padim, rudolph2022fully, deng2022anomaly, roth2022towards} utilize networks that are pre-trained on ImageNet for feature extraction. With a pre-trained model, PaDiM~\cite{defard2021padim} embeds the extracted anomaly patch features by multivariate Gaussian distribution. PatchCore~\cite{roth2022towards} uses a maximally representative memory bank of nominal patch features. Mahalanobis distance or maximum feature distance is adopted to score the input features in testing. However, industrial images generally have a different distribution from ImageNet. Directly using pre-trained features may cause a mismatch problem. Moreover, either computing the inverse of covariance~\cite{defard2021padim} or searching through the nearest neighbor in the memory bank~\cite{roth2022towards} limits the real-time performance, especially for edge devices.

CS-Flow~\cite{rudolph2022fully}, CFLOW-AD~\cite{gudovskiy2022cflow}, and DifferNet~\cite{rudolph2021same} propose to transform the normal feature distribution into Gaussian distribution via normalizing flow (NF)~\cite{rezende2015variational}. As normalizing flow can only process full-sized feature maps, i.e., down sample is not allowed and the coupling layer~\cite{dinh2016density} consumes a few times of memory than the normal convolutional layer, these methods are memory consuming. Distillation methods~\cite{bergmann2020uninformed, deng2022anomaly} train a student network to match the outputs of a fixed pre-trained teacher network with only normal samples. A discrepancy between student and teacher output should be detected given an anomalous query. 
The computational complexity is doubled as an input image should pass through both the teacher and the student.

SimpleNet overcomes the aforementioned problems.  SimpleNet uses a feature adaptor that performs transfer learning on the target dataset to alleviate the bias of pre-trained CNNs. SimpleNet proposes to synthesize anomalous in the feature space rather than directly on the images. SimpleNet follows a single-stream manner at inference and is totally constructed by conventional CNN blocks which facilitate fast training, inference, and industrial application. 

\section{Method}

The proposed SimpleNet is elaborately introduced in this section. As illustrated in Figure~\ref{Fig2}, SimpleNet consist of a \textit{Feature Extractor}, a \textit{Feature Adaptor}, an \textit{Anomalous Feature Generator} and a \textit{Discriminator}. 
The \textit{Anomalous Feature Generator} is only used during training, thus SimpleNet follows a single stream manner at inference. These modules will be described below in sequence.

\subsection{Feature Extractor}

Feature Extractor acquires local feature as in~\cite{roth2022towards}. We reformulate the process as follows. We denote the training set and test set as $\mathcal{X}_{train}$ and $\mathcal{X}_{test}$. For any image $x_{i}\in\mathbb{R}^{H\times W \times 3}$ in $\mathcal{X}_{train}\bigcup\mathcal{X}_{test}$ , the pre-trained network $\phi$ extracts features from different hierarchies, as normally done with ResNet-like backbone. Since pre-trained network is biased towards the dataset in which it is trained, it is reasonable to choose only a subset of levels for the target dataset. Formally, we define  $L$ the subset including the indexes of hierarchies for use. The feature map from level $l \in L$ is denoted as $\phi^{l, i} \sim  \phi^{l}(x_{i})\in\mathbb{R}^{H_{l}\times W_{l}\times C_{l}}$, where $H_{l}$, $W_{l}$ and $C_{l}$ are the height, width and channel size of the feature map. For an entry $\phi^{l,i}_{h,w}\in\mathbb{R}^{C_{l}}$ at location $(h, w)$, its neighborhood with patchsize $p$ is defined as 
\begin{equation} \label{eq1} \small
	\begin{split}
		\mathcal{N}_{p}^{(h,w)} =  \{ (h',y')| 
		& h'\in \left [ h-\left \lfloor  p/2\right \rfloor,...,h+\left \lfloor p/2 \right \rfloor  \right ], \\
		& y'\in \left [ w-\left \lfloor  p/2\right \rfloor,...,w+\left \lfloor p/2 \right \rfloor  \right ] \} 
	\end{split}
\end{equation}
Aggregating the features within the neighborhood $\mathcal{N}_{p}^{h,w}$ with aggregation function $f_{agg}$ (use adaptive average pooling here) results in the local feature $z_{h,w}^{l,i}$, as
\begin{equation} \label{eq2} \small
	z^{l,i}_{h,w} = f_{agg}(\{\phi^{l,i}_{h',y'}|(h',y')\in\mathcal{N}_{p}^{h,w}\})
\end{equation}
To combine features $z^{l,i}_{h,w}$ from different hierarchies, all feature maps are linearly resized to the same size $(H_{0}, W_{0})$, i.e. the size of the largest one. Simply concatenating the feature maps channel-wise gives the feature map $o^{i}\in\mathbb{R}^{H_{0}\times H_{0}\times C}$. The process is defined as
\begin{equation}\label{eq3} \small
	o^{i} = f_{cat}({resize(z^{l', i}, (H_{0}, W_{0}))|l'\in L}
\end{equation}
we define $o^{i}_{h, w} \in \mathbb{R}^{C}$ as the entry of $o^{i}$ at location $(h,w)$.

We simplify the above expressions as
\begin{equation} \small
	o^{i} = F_{\phi}(x^{i})
\end{equation}
where $F_{\phi}$ is the Feature Extractor.

\subsection{Feature Adaptor}

As industrial images generally have a different distribution from the dataset used in backbone pre-training, we adopt a Feature Adaptor $G_{\theta}$ to transfer the training features to the target domain. The Feature Adaptor $G_{\theta}$ projects local feature $o_{h, w}$ to adapted feature $q_{h, w}$ as 
\begin{equation} \label{eq6} \small
	q^{i}_{h, w}=G_{\theta}(o^{i}_{h,w})
\end{equation}

The Feature Adaptor can be made up of simple neural blocks such as a fully-connected layer or multi-layer perceptron (MLP). We experimentally find that a single fully-connected layer yields good performance.

\begin{figure}[t]
	\centering
	\includegraphics[width=0.45\textwidth] {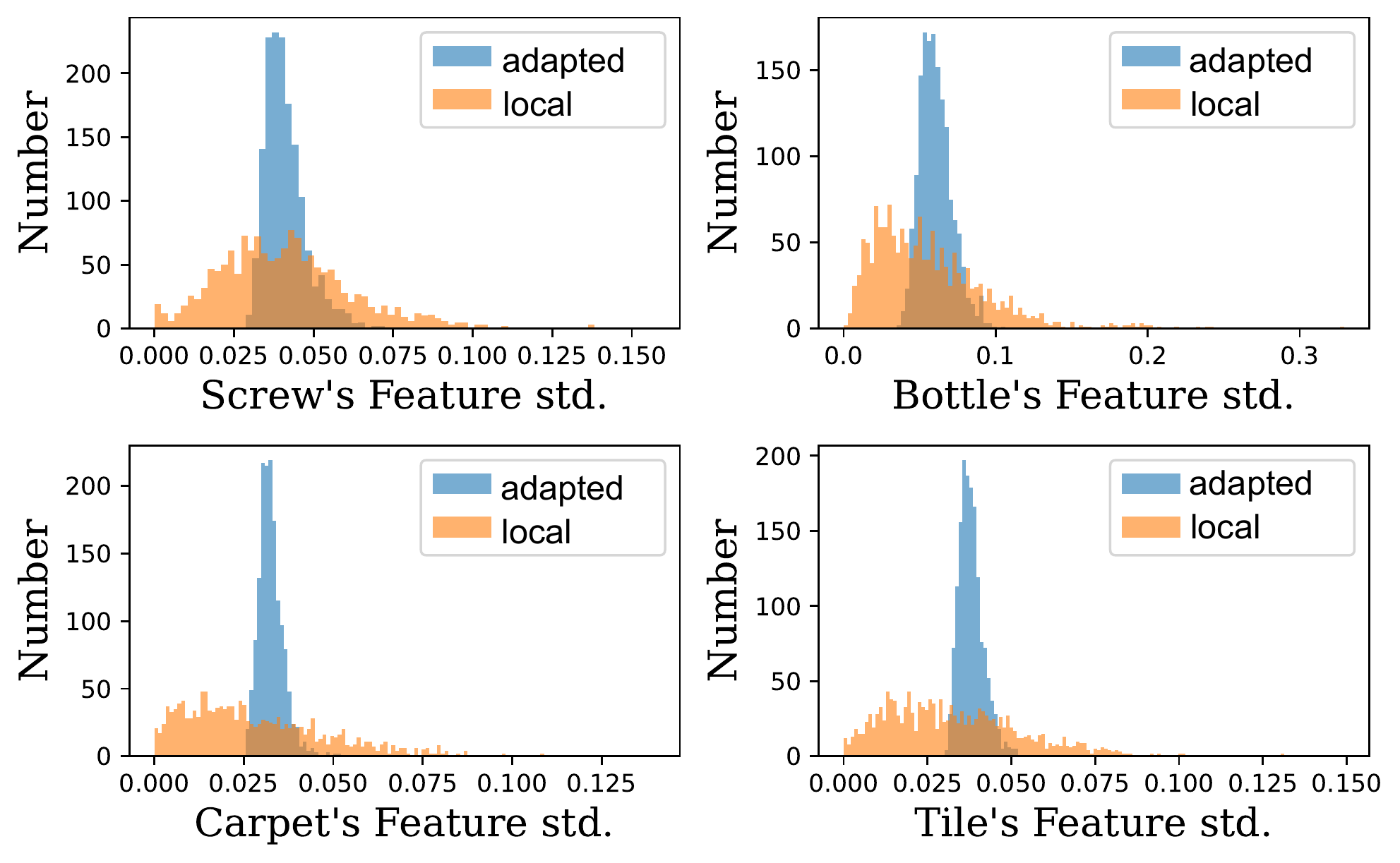}
	\caption{Histogram of standard deviation along each dimension of local feature and adapted feature. The adapted feature space becomes more compact when training with anomalous features.}
	\label{Fig:FA}
\end{figure}

\subsection{Anomalous Feature Generator}

To train the Discriminator to estimate the likelihood of samples being normal, the easiest way is sampling negative samples, i.e. defect features, and optimizing it together with normal samples. The lack of defects makes the sampling distribution estimation intractable. While~\cite{liznerskiexplainable,zavrtanik2021draem,li2021cutpaste} relying on extra data to synthesize defect images, we add simple noise on normal samples in the feature space, claiming that it outperforms those manipulated methods.

The anomalous features are generated by adding Gaussian noise on the normal features $q^{i}_{h, w}\in\mathbb{R}^{C}$. Formally, a noise vector $\epsilon\in\mathbb{R}^{C}$ is sampled, with each entry following an i.i.d. Gaussian distribution $\mathcal{N}(\mu, \sigma^{2})$. The anomalous feature $q^{i-}_{h, w}$ is fused as
\begin{equation} \small
	q^{i-}_{h, w} = q^{i}_{h, w} + \epsilon
\end{equation}

Figure~\ref{Fig:FA} illustrates the influence of anomalous features on four classes of MVTec AD. We can see that the standard deviation along each dimension of the adapted features tends to be consistent. Thus, the feature space tends to be compact when distinguishing anomalous features from normal features.

\subsection{Discriminator}
The Discriminator $D_{\psi}$ works as a normality scorer, estimating the normality at each location $(h, w)$ directly.
Since negative samples are generated along with normal features $\{q^{i}|x^{i}\in\mathcal{X}_{train}\}$, they are both fed to the Discriminator during training. The Discriminator expects positive output for normal features while negative for anomalous features. We simply use a 2-layer multi-layer perceptron (MLP) structure as common classifiers do, estimating normality as $D_{\psi}(q_{h, w})\in\mathbb{R}$.

\subsection{Loss function and Training}
\label{loss_func}
A simple truncated $l1$ loss is derived as 
\begin{equation} \label{eq7} \small
	l^{i}_{h,w} = \max(0, th^{+}-D_{\psi}(q^{i}_{h,w})) + \max(0, -th^{-}+D_{\psi}(q^{i-}_{h,w}))
\end{equation}
$th^{+}$ and $th^{-}$ are truncation terms preventing overfitting. They are set to 0.5 and -0.5 by default.
The training objective is 
\begin{equation} \label{eq9} \small
	\mathcal{L} = \min\limits_{\theta, \psi}\sum_{x^{i}\in\mathcal{X}_{train}}\sum_{h,w}\frac{{}l^{i}_{h,w}}{H_{0}*W_{0}}
\end{equation}
We will experimentally evaluate the proposed truncated $l1$ loss function with the widely used cross-entropy loss in the experiments section. 
The pseudo-code of the training procedure is shown in Algorithm~\ref{algo1}.

\subsection{Inference and Scoring function}
The Anomalous Feature Generator is discarded at inference. Note that the remaining modules can be stacked into an end-to-end network. We feed each $x_i \in \mathcal{X}_{test}$ into the aforementioned Feature Extractor $F_{\phi}$ and the Feature Adaptor $G_{\theta}$ sequentially to get adapted features $q^i_{h,w}$ as in Equation~\ref{eq6}. The anomaly score is provided by the Discriminator $D_{\psi}$ as
\begin{equation} \small
	s^{i}_{h, w} = -D_{\psi}(q^{i}_{h,w})
\end{equation}

The anomaly map for anomaly localization during inference is defined as
\begin{equation}  \label{eq18} \small
	S_{AL}(x_i):= \{s^{i}_{h, w}|(h,w)\in W_0\times H_0\}
\end{equation}

Then $S_{AL}(x_i)$ is interpolated to have the spatial resolution of the input sample and Gaussian filtered with $\sigma = 4$ for smooth boundaries. As the most responsive point exists for any size of the anomalous region, the maximum score of the anomaly map is taken as the anomaly detection score of each image
\begin{equation}  \label{eq19} \small
	S_{AD}(x_i):=\max_{(h,w)\in W_0\times H_0} s^i_{h,w}
\end{equation}

\section{Experiments}

\subsection{Datasets.} 
We conduct most of the experiments on the MVTec Anomaly Detection benchmark~\cite{bergmann2019mvtec}, that is, a famous dataset in the anomaly detection and localization field. MVTec AD contains 5 texture and 10 object categories stemming from manufacturing with a total of $5354$ images.
The dataset is composed of normal images for training and both normal and anomaly images with various types of defect for test. 
It also provides pixel-level annotations for defective test images. Typical images are illustrated in Figure~\ref{Fig1}. As in~\cite{defard2021padim,roth2022towards}, images are resized and center cropped to $256 \times 256$ and $224 \times 224$, respectively. No data augmentation is applied. We follow the one-class classification protocol, also known as cold-start anomaly detection, where we train a one-class classifier for each category on its respective normal training samples.

We conduct one-class novelty detection on CIFAR10~\cite{krizhevsky2009learning}, which contains 50K training images and 10K test images with scale of $32 \times 32$ in 10 categories. Under the setting of one-class novelty detection, one category is regarded as normal data and other categories are used as novelty.

\definecolor{dkgreen}{rgb}{0,0.6,0}
\definecolor{gray}{rgb}{0.5,0,0}
\definecolor{mauve}{rgb}{0.58,0,0.82}

\lstset{
	language=Python,
	basicstyle=\small\ttfamily\bfseries,
	numberstyle=\color{gray},
	xleftmargin=5pt,
	framexleftmargin=5pt,
	commentstyle=\color{dkgreen},
	keywordstyle=\color{blue},
}

\begin{algorithm}[t]
\setstretch{0.85} 
\caption{SimpleNet training pseudo-code, Pytorch-like}
\begin{lstlisting}  
# F: Feature Extractor
# G: Feature Adaptor
# N: i.i.d Gaussian noise
# D: Discriminator
pretrain_init(F)
random_init(G, D)
for x in data_loader:
    o = F(x)  # normal features
    q = G(o)  # adapted features
    q_ = q + random(N) # anomalous features
	
    loss = loss_func(D(q), D(q_)).mean()
    loss.backward() # back-propagate

    F = F.detach() # stop gradient
    update(G, D) # Adam

# loss function    
def loss_func(s, s_):
    th_ = -th = 0.5
    return max(0, th-s) + max(0, th_+s_)
\end{lstlisting}
\label{algo1}
\end{algorithm}

\subsection{Evaluation Metrics.} Image-level anomaly detection performance is measured via the standard Area Under the Receiver Operator Curve, which we denote as I-AUROC, using produced anomaly detection scores $S_{AD}$ (Equation~\ref{eq19}). For anomaly localization, the anomaly map $S_{AL}$ (Equation~\ref{eq18}) is used for an evaluation of pixel-wise AUROC (denoted as P-AUROC). In accordance with prior works~\cite{defard2021padim,roth2022towards}, we compute on MVTec AD the class-average AUROC and mean AUROC overall categories for detection and localization. 
The comparison baselines includes AE-SSIM~\cite{bergmann2019mvtec}, RIAD~\cite{zavrtanik2021reconstruction}, DRÆM~\cite{zavrtanik2021draem}, CutPaste~\cite{li2021cutpaste}, CS-Flow~\cite{rudolph2022fully}, PaDiM~\cite{defard2021padim}, RevDist~\cite{deng2022anomaly} and PatchCore~\cite{roth2022towards}.

\begin{table*}[t]
	\footnotesize
	\caption{Comparison of SimpleNet with state-of-the-arts works on MVTec AD. Image-wise AUROC (I-AUROC) and pixel-wise AUROC (P-AUROC) are displayed in each entry as I-AUROC\%/P-AUROC\%. P-AUROC for CS-Flow is not recorded in \cite{rudolph2022fully}}
	
	\centering 
	\begin{tabular}{|l|c|c|c|c|c|c|c|c|c|c|}
		\hline
		\multicolumn{1}{|l|}{Type} & \multicolumn{2}{c|}{Reconstruction-based} & \multicolumn{2}{c|}{Synthesizing-based}                    & \multicolumn{4}{c|}{Embedding-based}                                                  & \multicolumn{1}{c|}{Ours}   \\ 
		\hline
		Model                      & AE-SSIM & RIAD                            & DRÆM                       & CutPaste                   & CS-Flow        & PaDiM              & RevDist                    & PatchCore          & SimpleNet                   \\
		\hline
		Carpet     & 87/64.7 & 84.2/96.3                      & 97.0/95.5                  & 93.9/98.3                  & \textbf{100}/-  & 99.8/99.1         & 98.9/98.9                  & 98.7/\textbf{99.0} & 99.7/98.2                      \\
		Grid       & 94/84.9 & 99.6/98.8                      & 99.9/99.7                  & \textbf{100}/97.5          & 99.0/-          & 96.7/97.3         & \textbf{100}/\textbf{99.3} & 98.2/98.7          & 99.7/98.8                      \\
		Leather    & 78/56.1 & \textbf{100}/99.4              & \textbf{100}/98.6          & \textbf{100}/\textbf{99.5} & \textbf{100}/-  & \textbf{100}/99.2 & \textbf{100}/99.4          & \textbf{100}/99.3  & \textbf{100}/99.2              \\
		Tile       & 59/17.5 & 98.7/89.1                      & 99.6/\textbf{99.2}         & 94.6/90.5                  & \textbf{100}/-  & 98.1/94.1         & 99.3/95.6                  & 98.7/95.6          & 99.8/97.0                      \\
		Wood       & 73/60.3 & 93.0/85.8                      & 99.1/\textbf{96.4}         & 99.1/95.5                  & \textbf{100}/-  & 99.2/94.9         & 99.2/95.3                  & 99.2/95.0          & \textbf{100}/94.5             \\
		\hline
		Avg. Text. & 78/56.7 & 95.1/93.9                      & 99.1/\textbf{97.9}         & 97.5/96.3                  & \textbf{99.8}/- & 95.5/96.9         & 99.5/97.7                  & 99.0/97.5          & \textbf{99.8}/97.5            \\
		\hline
		Bottle     & 93/83.4 & 99.9/98.4                      & 99.2/\textbf{99.1}         & 98.2/97.6                  & 99.8/-          & 99.1/98.3         & \textbf{100}/98.7          & \textbf{100}/98.6  & \textbf{100}/98.0              \\
		Cable      & 82/47.8 & 81.9/84.2                      & 91.8/94.7                  & 81.2/90.0                  & 99.1/-          & 97.1/96.7         & 95.0/97.4                  & 99.5/\textbf{98.4} & \textbf{99.9}/97.6            \\
		Capsule    & 94/86.0 & 88.4/92.8                      & \textbf{98.5}/94.3         & 98.2/97.4                  & 97.1/-          & 87.5/98.5         & 96.3/98.7                  & 98.1/98.8          & 97.7/\textbf{98.9}            \\
		Hazelhut   & 97/91.6 & 83.3/96.1                      & \textbf{100}/\textbf{99.7} & 98.3/97.3                  & 99.6/-          & 99.4/98.2         & 99.9/98.9                  & \textbf{100}/98.7  & \textbf{100}/97.9            \\
		Metal Nut  & 89/60.3 & 88.5/92.5                      & 98.7/\textbf{99.5}         & 99.9/93.1                  & 99.1/-          & 96.2/97.2         & \textbf{100}/97.3          & \textbf{100}/98.4  & \textbf{100}/98.8              \\
		Pill       & 91/83.0 & 83.8/95.7                      & 98.9/97.6                  & 94.9/95.7                  & 98.6/-          & 90.1/95.7         & 96.6/98.2                  & 96.6/97.4          & \textbf{99.0}/\textbf{98.6}   \\
		Screw      & 96/88.7 & 84.5/98.8                      & 93.9/97.6                  & 88.7/96.7                  & 97.6/-          & 97.5/98.5         & 97.0/\textbf{99.6}         & 98.1/99.4          & \textbf{98.2}/99.3            \\
		Toothbrush & 92/78.4 & \textbf{100}/98.9              & \textbf{100}/98.1          & 99.4/98.1                  & 91.9/-          & \textbf{100}/98.8 & 99.5/\textbf{99.1}         & \textbf{100}/98.7  & 99.7/98.5                     \\
		Transistor & 90/72.5 & 90.9/87.7                      & 93.1/90.9                  & 96.1/93.0                  & 99.3/-          & 94.4/97.5         & 96.7/92.5                  & \textbf{100}/96.3  & \textbf{100}/\textbf{97.6}    \\
		Zipper     & 88/66.5 & 98.1/97.8                      & \textbf{100}/98.8          & 99.9/99.3                  & 99.7/-          & 98.6/98.5         & 98.5/98.2                  & 99.4/98.8          & 99.9/\textbf{98.9}             \\
		\hline
		Avg. Obj.  & 91/75.8 & 89.9/94.3                      & 97.4/97.0                  & 95.5/95.8                  & 98.2/-          & 96.0/97.8         & 98/97.9                    & 99.2/\textbf{98.4} & \textbf{99.5}/\textbf{98.4}    \\
		\hline
		Average    & 87/69.4 & 91.7/94.2                      & 98.0/97.3                  & 96.1/96.0                  & 98.7/-          & 95.8/97.5         & 98.5/97.8                  & 99.1/\textbf{98.1} & \textbf{99.6}/\textbf{98.1}   \\ \hline                   
	\end{tabular}
	\label{table:1}
\end{table*}

\begin{table}\caption{Performance on MVTec AD under different combinations of hierarchy levels of WideResNet50 to use.}\centering
	\footnotesize
	\begin{tabular}{|c|c|c|c|c|}
		\hline
		level1     & level2    & level3    & I-AUROC\%     & P-AUROC\% \\
		\hline
		\checkmark &           &           & 93.0          & 94.2      \\
		& \checkmark&           & 98.4          & 96.7      \\
		&           & \checkmark& 99.2          & 97.5      \\
		\checkmark & \checkmark&           & 96.7          & 96.7      \\
		& \checkmark&\checkmark & \textbf{99.6} & \textbf{98.1}     \\
		\checkmark & \checkmark& \checkmark& 99.1          & \textbf{98.1}     \\
		\hline 
	\end{tabular}
	\label{table:20}
\end{table}

\subsection{Implementation Details}
This section describes the configuration implementation details of the experiments in this paper. All backbones used in the experiments were pre-trained with ImageNet~\cite{deng2009imagenet}. The 2nd and 3rd intermediate layers of the backbone e.g. $l' \in [2,3]$ in Equation~\ref{eq3} are used in the feature extractor as in~\cite{roth2022towards} when the backbone is ResNet-like architecture. By default, our implementation uses WideResnet50 as backbone, and the feature dimension from the feature extractor is set to 1536. 
The later feature adaptor is essentially a fully connected layer without bias. The dimensions of the input and output features for the FC layer in the adaptor are the same. The anomaly feature generator adds i.i.d. Gaussian noise $\mathcal{N}(0, \sigma^{2})$ to each entry of normal features. $\sigma $ is set to 0.015 by default. 
The subsequent discriminator composes of a linear layer, a batch normalization layer, a leaky relu(0.2 slope), and a linear layer. $th^{+}$ and $th^{-}$ are both set to $0.5$ in Equation~\ref{eq7}. The Adam optimizer is used, setting the learning rate for the feature adaptor and discriminator to 0.0001 and 0.0002 respectively, and weight decay to 0.00001. Training epochs is set to 160 for each dataset and batchsize is 4.

\subsection{Anomaly detection on MVTec AD}

Anomaly detection results on MVTec AD are shown in Table~\ref{table:1}. Image-level anomaly score is given by the maximum score of the anomaly map as in Equation~\ref{eq19}. SimpleNet achieves the highest score for 9 out of 15 classes. For textures and objects, SimpleNet reaches new SOTA of $99.8\%$ and $99.5\%$ of I-AUROC, respectively. SimpleNet achieves significantly higher mean image anomaly detection performance i.e. I-AUROC score of $99.6\%$. Please note that, a reduction from an error of $0.9\%$ for PatchCore~\cite{roth2022towards} (next best competitor, under the same WideResnet50 backbone) to $0.4\%$ for SimpleNet means a reduction of the error by $55.5\%$. In industrial inspection settings, this is a relevant and significant reduction.

\begin{figure*}[t]
	\centering
	\includegraphics[width=0.4\textwidth]{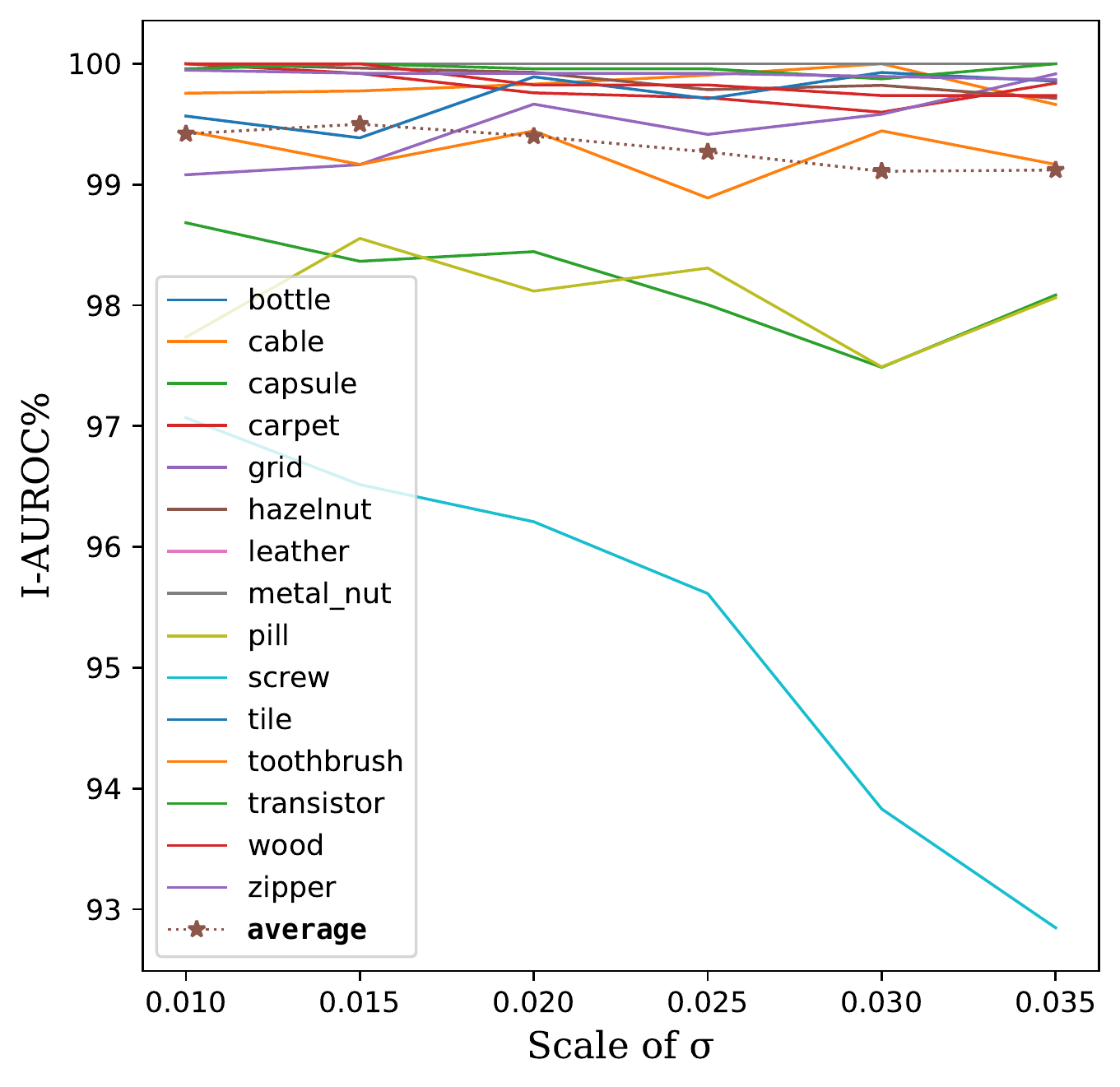}
	\hspace{0.5in}
	\includegraphics[width=0.4\textwidth]{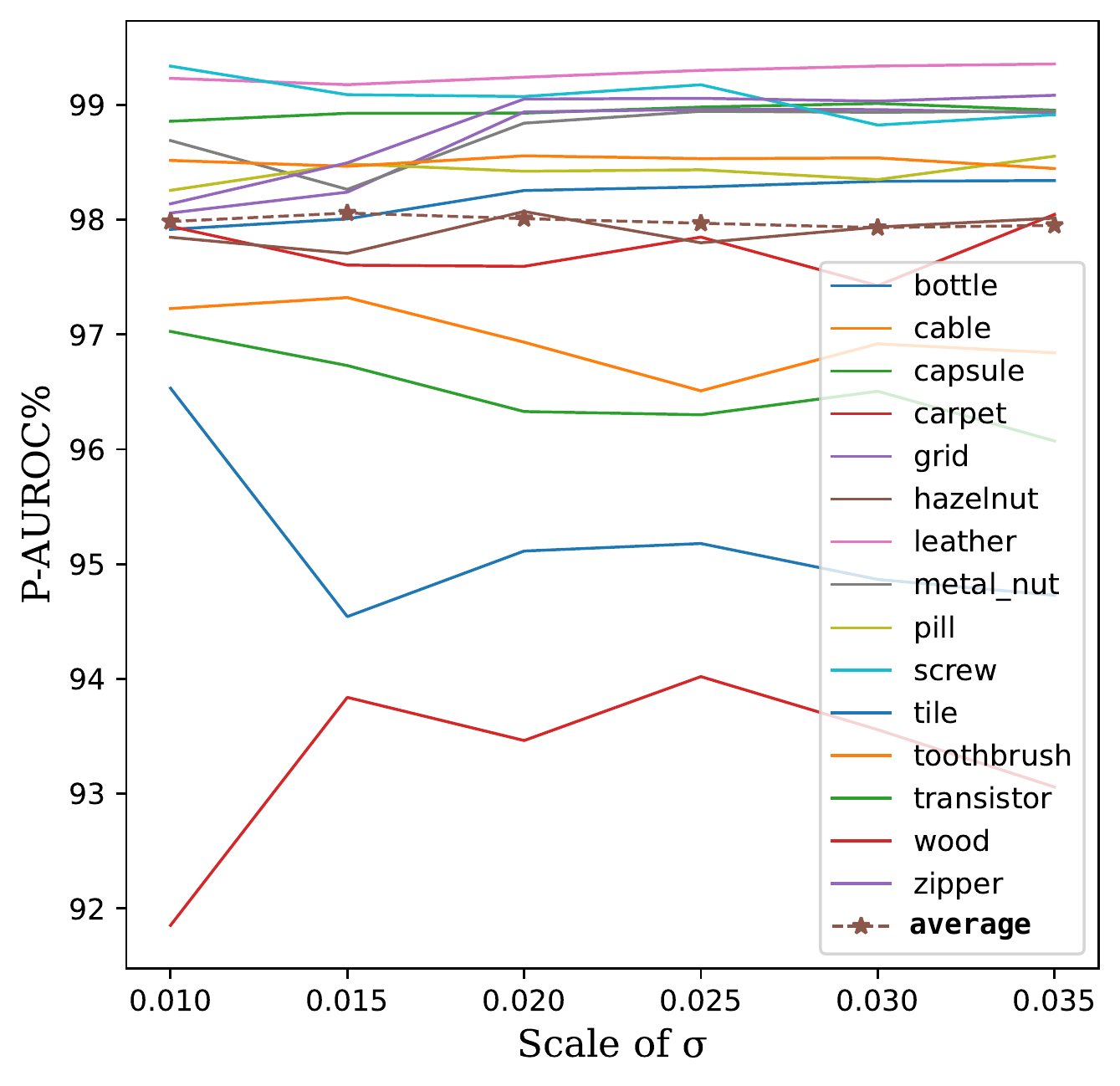}
	\caption{I-AUROC\%  and P-AUROC\% for each class of MVTec AD dataset with varied $\sigma$. (Best viewed in color.)}
	\label{Fig4}
\end{figure*}

\subsection{Anomaly localization on MVTec AD}
The anomaly localization performance is measured by pixel-wise AUROC, which we note as P-AUROC. Comparisons with the state-of-the-art methods are shown in Table~\ref{table:1}. SimpleNet achieves the best anomaly detection performance of $98.1\%$ P-AUROC on MVTec AD as well as the new SOTA of $98.4\%$ P-AUROC for objects. SimpleNet achieves the highest score for 4 out of 15 classes. We visualize representative samples for anomaly localization in Figure~\ref{Fig8}.

\subsection{Inference time}
Alongside the detection and localization performance, inference time is the most important concern for industrial model deployment. The comparison with the state-of-the-art methods on inference time is shown in Figure~\ref{Fig:speed}. All the methods are measured on the same hardware containing a Nvidia GeForce GTX 3080ti GPU and an Intel(R) Xeon(R) CPU E5-2680 v3@2.5GHZ. It clearly shows that our method achieves the best performance as well as the fastest speed at the same time. SimpleNet is nearly 8$\times$ faster than PatchCore~\cite{roth2022towards}.

\subsection{Ablation study}
\label{Ablation}

\begin{figure}[t]
	\centering
	\includegraphics[width=0.4\textwidth]{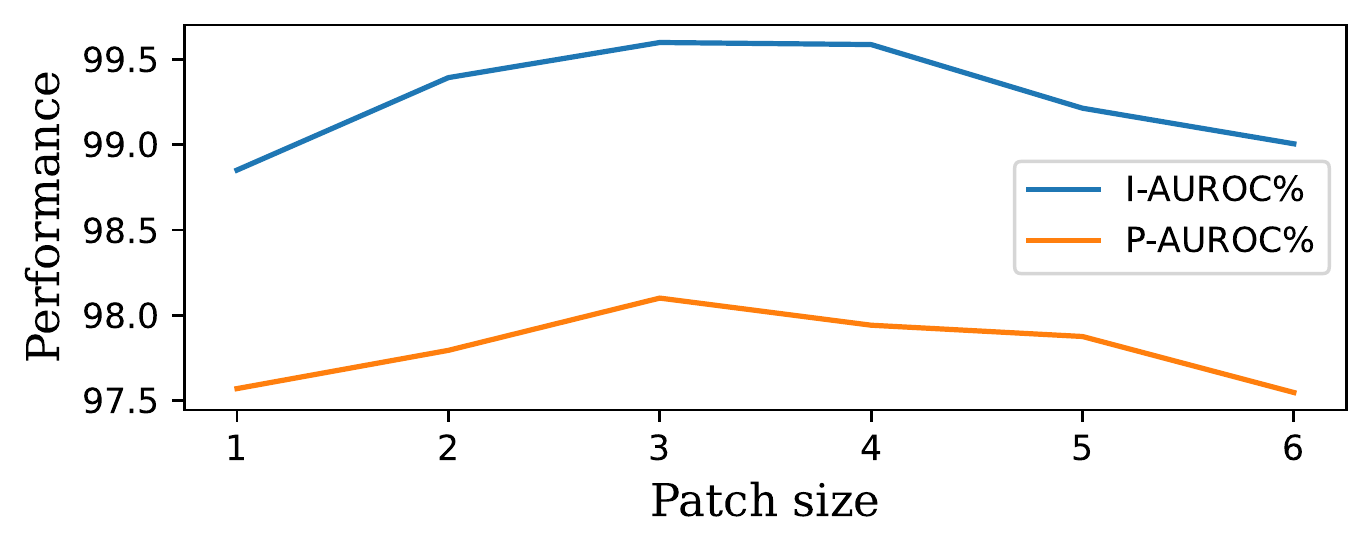} 
	\caption{Performance with varied patch sizes on MVTec AD.}
	\label{Fig9}
\end{figure}

\textbf{Neighborhood size and hierarchies.} We investigate the influence of neighborhood size $p$ in Equation~\ref{eq1}. 
Results in Figure~\ref{Fig9} show a clear optimum between locality and global context for anomaly predictions, thus motivating the neighborhood size $p =3$. 
We design a group of experiments to test the influence of hierarchies subset $L$ on model performance and the results are shown in Table~\ref{table:20}. We index the first three WideResNet50 blocks with $1-3$. As can be seen, features from hierarchy level 3 can already achieve state-of-the-art performance but benefit from additional hierarchy level 2. We chose $2+3$ as the default setting.

\begin{figure}[t]
	\centering
	\includegraphics[width=0.4\textwidth]{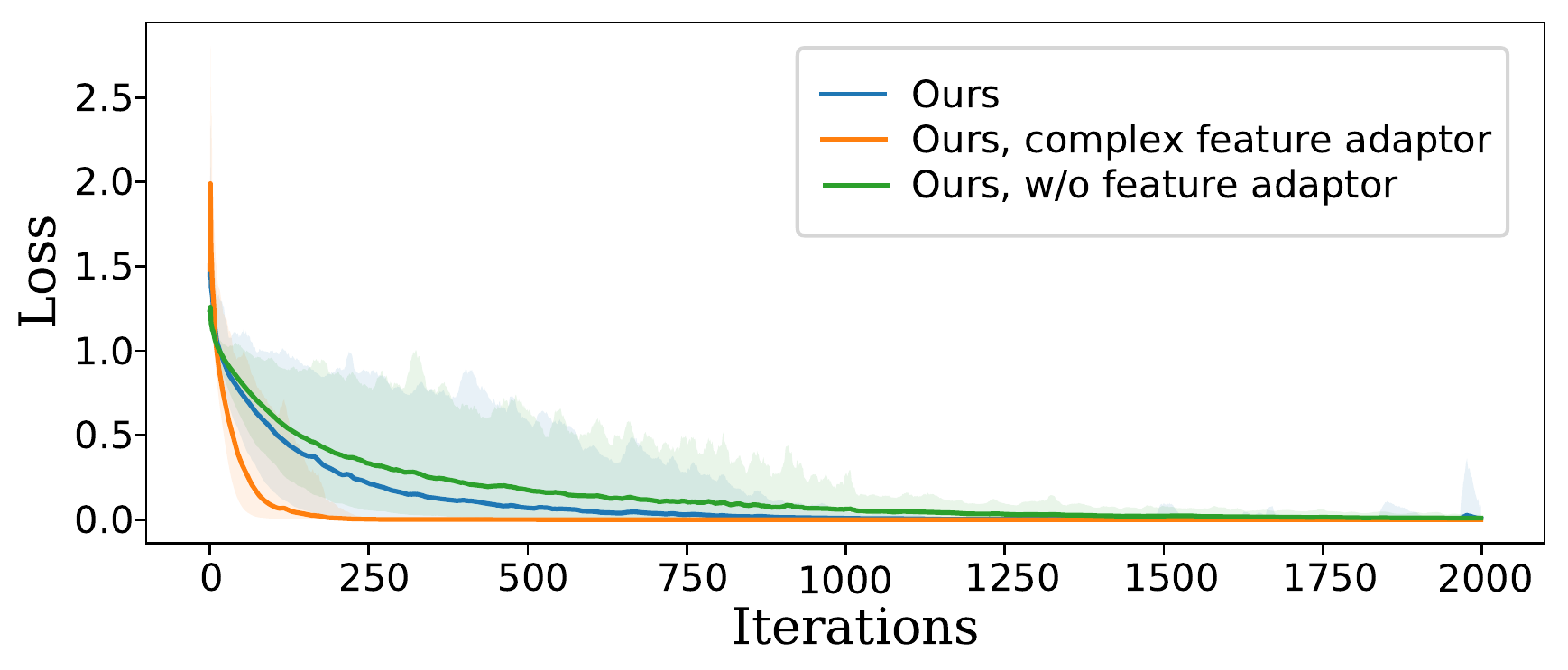}
	\caption{Visualization of loss during training. The plotted lines show the mean loss for all classes in the MVTec AD dataset. The transparent color shows the range of loss fluctuation.}
	\label{Fig6}
\end{figure}

\begin{figure*}[t]
	\centering
	\includegraphics[width=0.95\textwidth]{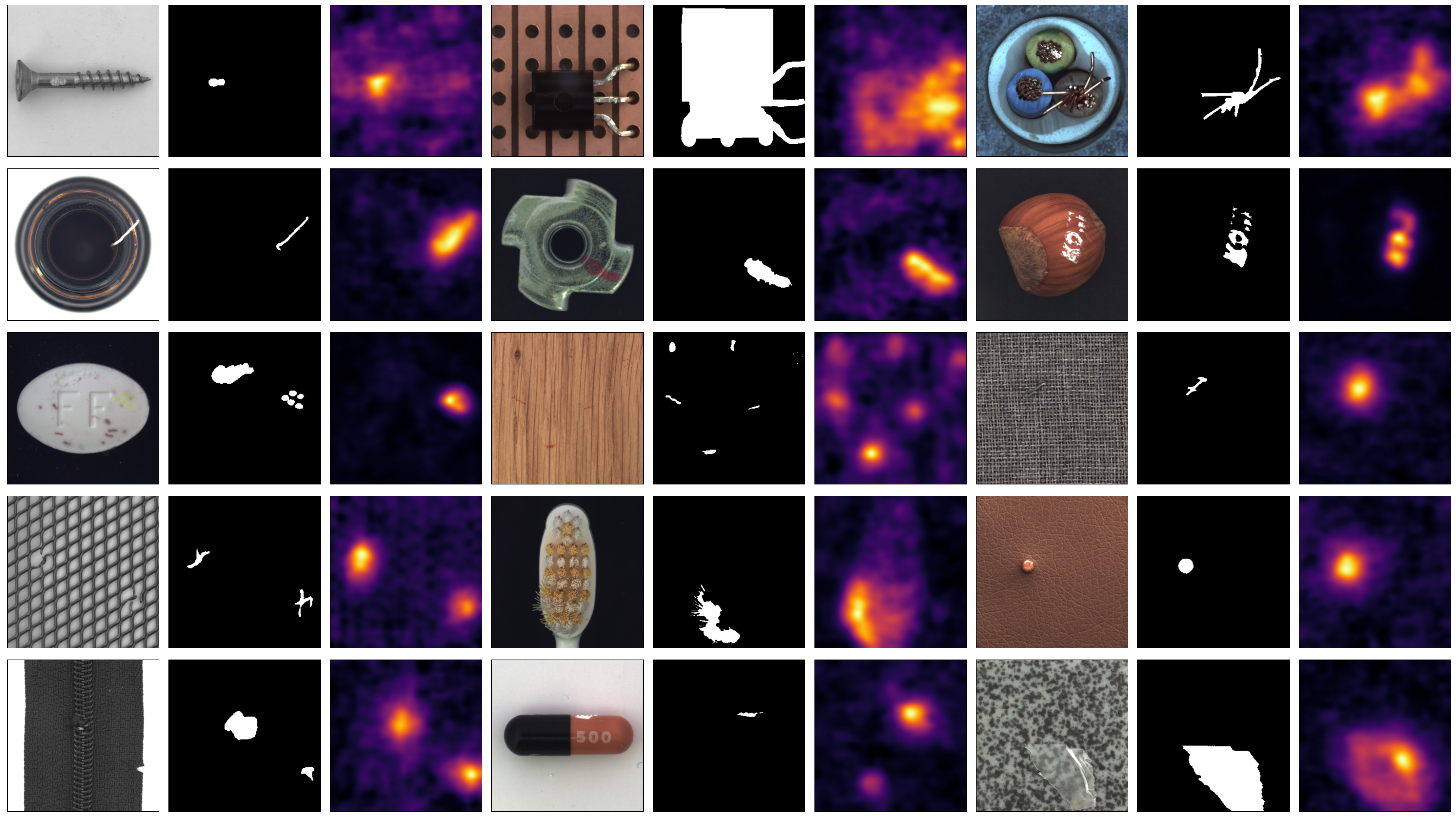} 
	\caption{Qualitative results, where sampled image, ground truth, and anomaly map are shown for each class in MVTec AD.}
	\label{Fig8}
\end{figure*}

\textbf{Adaptor configuration.} Adaptor provides a transformation (projection) on the pre-trained features. Our default feature adaptor is a single FC layer without bias, with equal input and output channels. A comparison of different feature adaptors is shown in Table~\ref{table:2}, the first row ”Ours” implementation follows the same configuration as in Table~\ref{table:1}. “Ours-complex-FA” replaces the simple feature adaptor with a nonlinear one (i.e. 1 layer MLPs with nonlinearity). The row ”Ours-w/o-FA” drops the feature adaptor. The results indicate that a single FC layer yields the best performance. Intuitively, the feature adaptor finds a projection such that the faked abnormal features and projected pre-trained features are easily severed, meaning a simple solution to the discriminator. This is also indicated by the phenomenon that using a feature adaptor helps the network converge fast (Figure~\ref{Fig6}). We observe a significant performance drop with a complex feature adaptor. One possible reason is that a complex adaptor may lead to overfitting, reducing the generalization ability for various defects in test. 
Figure~\ref{Fig:FA} compares the histogram of standard deviation along each dimension of the features before and after the feature adaptor. We can see that, when training with anomalous features, adapted feature space becomes compact.

\begin{table}\footnotesize\caption{Comparison of different feature adaptors. "Ours" implementation follows the same configuration as in Table \ref{table:1}. "Ours-complex-FA" replaces the simple feature adaptor with a nonlinear one. "Ours-w/o-FA" drops the feature adaptor, equivalent to using an identity fully-connected layer. "Ours-CE" uses cross-entropy loss. I-AUROC\% and P-AUROC\% of MVTec AD are shown.}\centering

	\begin{tabular}{|c|c|c|}
		\hline
		Model & I-AUROC\% & P-AUROC\% \\
		\hline
		Ours & \textbf{99.6} & \textbf{98.1} \\
		\hline
		Ours-complex-FA & 98.3 & 97.2 \\
		\hline
		Ours-w/o-FA & 99.2 & 97.9 \\
		\hline
		Ours-CE & 99.4 & 97.8 \\ 
		\hline
	\end{tabular}
	\label{table:2}
\end{table}

\begin{table}
	\centering
	\caption{Performance under different backbones on MVTec AD. }\centering
	\footnotesize
	\begin{tabular}{|c|c|c|} 
		\hline
		Model           & I-AUROC\% & P-AUROC\%  \\ 
		\hline
		ResNet18        & 98.3    & 95.7     \\ 
		\hline
		ResNet50        & 99.6    & 98.0       \\ 
		\hline
		ResNet101   & 99.2    & 97.6     \\ 
		\hline
		WideResNet50    & \textbf{99.6}    & \textbf{98.1}\\
		\hline
	\end{tabular}
	\label{table:backbone}
\end{table}

\begin{table} 
	\caption{One-Class Novelty Detection I-AUROC(\%) results on CIFAR-10 dataset.}
	\centering
	\small
	\resizebox{1\columnwidth}{!}{
		\begin{tabular}{|l|c|c|c|c|c|} 
			\hline
			Method & LSA         & DSVDD    & OCGAN & HRN     & DAAD                   \\ 
			\hline
			AUROC  & 64.1        & 64.8     & 65.6  & 71.3    & 75.3      \\ 
			\hline
			Method & DisAug CLR~ & IGD      & MKD   & RevDist & \textbf{\textbf{SimpleNet}} \\ 
			\hline
			AUROC  & 80.0        & 83.68 & 84.5  & 86.5    & \textbf{86.5}     \\
			\hline
		\end{tabular}
	}
	\label{one_class}
\end{table}

\textbf{Scale of noise.} The scale of noise in the anomaly feature generator controls how far away the synthesized abnormal features are from the normal ones. To be specific, high $\sigma$ results in abnormal features keeping a high Euclidean distance towards normal features. Training on a large $\sigma$ will result in a loose decision bound, leading to a high false negative. Conversely, the training procedure will become unstable if $\sigma$ is tiny, and the discriminator cannot generalize to normal features well. Figure~\ref{Fig4} details the effect of $\sigma$ for each class in MVTec AD. As can be seen, $\sigma = 0.015$ reaches the balance and yield the best performance. 

\textbf{Loss function.} We compared the proposed loss function in Section~\ref{loss_func} with the widely used cross-entropy loss (as show in row "Ours-CE" in Table~\ref{table:2}). We found the improvements, $0.2 \%$ I-AUROC and $0.3 \%$ P-AUROC, over cross-entropy loss. 

\textbf{Dependency on backbone.} We test SimpleNet with different backbones, the results are shown in Table~\ref{table:backbone}. We find that results are mostly stable over the choice of different backbones. 
The choice of WideResNet50 is made to be comparable with PaDiM~\cite{defard2021padim} and PatchCore~\cite{roth2022towards}.

\textbf{Qualitative Results} Figure~\ref{Fig8} shows results of anomaly localization that indicate the abnormal areas. The threshold for segmentation results is obtained by calculating the F1-score for all anomaly scores of each sub-class. Experimental results prove that the proposed method can localize abnormal areas well even in rather difficult cases. In addition, we can find that the proposed method has consistent performance in both object and texture classes.

\subsection{One-Class Novelty Detection}
To evaluate the generality of the proposed SimpleNet, we conduct a one-class novelty detection experiment on CIFAR-10~\cite{krizhevsky2009learning}. Following ~\cite{perera2019ocgan}, we train the model with samples from a single class and detect novel samples from other categories. We train the corresponding model for each class respectively. Note that the novelty score is defined as the max score in the similarity map. Table~\ref{one_class} reports the I-AUROC scores of our method and other methods. For fair comparison, all the methods are pre-trained on ImageNet. The baselines include VAE~\cite{an2015variational}, LSA~\cite{abati2019latent}, DSVDD~\cite{ruff2018deep}, OCGAN~\cite{perera2019ocgan}, HRN~\cite{hu2020hrn}, AnoGAN~\cite{schlegl2017unsupervised}, DAAD~\cite{hou2021divide}, MKD~\cite{salehi2021multiresolution}, DisAug CLR~\cite{sohnlearning2021}, IGD~\cite{chen2022deep}
and RevDist~\cite{deng2022anomaly}. Our method outperforms these comparison methods. Note that, IGD~\cite{chen2022deep} and DisAug CLR~\cite{sohnlearning2021} achieve $91.25 \%$  and $92.4 \%$ respectively when boosted by self-supervised learning. 

\section{Conclusion}
In this paper, we propose a simple but efficient approach named SimpleNet for unsupervised anomaly detection and localization. SimpleNet consists of several simple neural network modules which are easy to train and apply in industrial scenarios. Though simple, SimpleNet achieves the highest performance as well as the fastest inference speed compared to the previous state-of-the-art methods 
on the MVtec AD benchmark. SimpleNet provides a new perspective to bridge the gap between academic research and industrial application in anomaly detection and localization.

\section*{Acknowledgments}
This work is supported by the National Natural Science Foundation of China under Grant 62176246 and Grant 61836008. This work is also supported by Anhui Provincial Natural Science Foundation 2208085UD17 and the Fundamental Research Funds for the Central Universities (WK3490000006).

{\small
\bibliographystyle{ieee_fullname}
\bibliography{egbib}
}

\end{document}